\documentclass[11pt,a4paper]{article}

\usepackage[utf8]{inputenc}
\usepackage[T1]{fontenc}
\usepackage{amsmath,amssymb}
\usepackage{graphicx}
\usepackage{booktabs}
\usepackage{multirow}
\usepackage{array}
\usepackage{url}
\usepackage{hyperref}
\usepackage{xcolor}
\usepackage{natbib}
\usepackage{geometry}
\usepackage{caption}
\usepackage{subcaption}
\usepackage{float}
\usepackage{tabularx}

\hypersetup{
  colorlinks=true,
  linkcolor=blue!70!black,
  citecolor=blue!70!black,
  urlcolor=blue!70!black
}

\title{
  \textbf{Pharmacogenomic Knowledge Graph Augmentation for \\
  Graph Neural Network-Based Drug-Drug Interaction Prediction}
}

\author{
  J\"{u}rgen Dietrich \\
  \textit{AI Solutions Berlin} \\
  \texttt{juergen.dietrich@ai-solutions-berlin.de} \\
  ORCID: 0000-0002-5494-3499
}

\date{}

\begin{document}

\maketitle

\begin{abstract}
Graph neural networks (GNNs) applied to drug-drug interaction (DDI) prediction
rely exclusively on molecular structure encoded as SMILES-derived graphs.
Prior work in this series demonstrated that model performance is bounded by
the structural information content of training labels---an
\emph{Information Ceiling}---that architectural refinements alone cannot
overcome. The present study investigates whether pharmacogenomic prior
knowledge from the PharmGKB database partially closes this ceiling by
providing metabolic pathway context that is independent of, and complementary
to, molecular structure.

Cytochrome P450 (CYP) enzyme substrate, inhibitor, and inducer annotations
for four clinically relevant isoforms (CYP2D6, CYP3A4, CYP2C19, CYP2C9) are
extracted and incorporated as a 12-dimensional feature vector concatenated to
the molecular embedding prior to interaction prediction. Experiments are
conducted under both pair-level and drug-level data splits to quantify
generalization to unseen drugs.

Results indicate that knowledge graph (KG) augmentation substantially improves
DDI type classification under pair-level split conditions (F1-macro: 0.532
vs.\ 0.241 baseline), while binary interaction detection and drug-level
generalization remain bounded by the Information Ceiling (area under the
receiver operating characteristic curve (AUC) inflation: 0.224 vs.\ 0.250
baseline). Mechanistic validation on strictly held-out compounds confirms
that augmentation preferentially improves CYP2C9-mediated interaction
prediction, with probabilities increasing from 0.033--0.117 (baseline)
to 0.560--0.586 (KG-augmented). An extension to single-molecule toxicity prediction on the Tox21
benchmark confirms that the effect is contingent on pharmacogenomic annotation
coverage. These findings motivate the multimodal framework proposed for the
subsequent study in this series.

\bigskip
\noindent\textbf{Keywords:} drug-drug interaction prediction, graph neural
networks, knowledge graph augmentation, pharmacogenomics, cytochrome P450,
information ceiling, molecular representation learning, toxicity prediction, Tox21
\end{abstract}

\newpage

\section{Introduction}

Drug-drug interactions (DDIs) constitute a significant source of adverse drug
events, affecting an estimated 20--30\% of hospitalized patients receiving
polypharmacy regimens~\cite{ddi_prevalence}. Accurate computational prediction
of DDIs from molecular structure alone has been the subject of intensive
research, with graph neural network (GNN) architectures demonstrating
state-of-the-art performance on benchmark datasets~\cite{ssiddi,pgnn}.

A preceding study in this series~\cite{paperC} identified a fundamental
limitation of structure-based DDI prediction, formalized as the
\emph{Information Ceiling Hypothesis} (H4): model performance is bounded by
the structural information content of training labels, $I(S, L)$, rather than
by architectural complexity. Empirical evidence for H4 was obtained by
comparing pair-level and drug-level data splits: under drug-level evaluation,
where test drugs are structurally unseen during training, AUC decreases by
approximately 0.25 units relative to pair-level benchmarks. This gap persists
irrespective of architectural refinements, including physical edge
construction~\cite{paperC}, confirming that the ceiling is determined by label
information rather than model capacity.

The present study addresses a natural extension: can pharmacogenomic prior
knowledge partially close this ceiling by providing metabolic pathway context
that is independent of, and complementary to, molecular structure? CYP enzymes
metabolize approximately 75\% of clinically used drugs~\cite{cyp_review}, and
their substrate, inhibitor, and inducer relationships---curated in the
PharmGKB database~\cite{pharmgkb}---constitute a rich source of mechanistic
information not captured by SMILES-derived molecular graphs.

The central hypothesis is that explicitly encoding CYP enzyme interaction
roles as molecular prior knowledge improves DDI prediction, particularly
for interaction types mediated by CYP-dependent pharmacokinetic mechanisms.

\subsection{Contributions}

The present study makes the following contributions:

\begin{enumerate}
  \item A pharmacogenomic feature extraction pipeline integrating three
        PharmGKB data sources (clinical variant annotations, guideline
        free-text, and relationship annotations) into a unified 12-dimensional
        CYP feature vector per drug molecule.

  \item A knowledge graph-augmented GNN architecture extending the physical
        edge model from~\cite{paperC} with a learned projection of the combined
        molecular embedding and CYP prior knowledge.

  \item Empirical evaluation under both pair-level and drug-level splits,
        with direct comparison to structure-only baselines, including
        quantification of AUC inflation under each split strategy.

  \item An extension to multi-task toxicity prediction on the Tox21
        benchmark~\cite{tox21}, investigating the coverage-dependence of the
        KG augmentation effect.
\end{enumerate}

\section{Related Work}

\subsection{Graph Neural Network-Based DDI Prediction}

Graph neural networks have been applied to DDI prediction through various
formulations. The SSI-DDI framework~\cite{ssiddi} is based on
substructure-substructure interactions, achieving strong performance on the
DrugBank benchmark. Physical interaction edges derived from hydrogen bonding
and electrostatic complementarity were investigated in~\cite{paperC}, yielding
a 51\% improvement in macro-averaged F1 score for DDI type classification.

\subsection{Knowledge Graph Augmentation}

Knowledge graph-based approaches to DDI prediction have predominantly focused
on entity embeddings derived from biomedical knowledge graphs such as
DRKG~\cite{drkg} and PharmKG~\cite{pharmkg}. These approaches typically
represent drugs as nodes in a heterogeneous graph and propagate information
through relational paths, yielding latent embeddings of 128--512 dimensions.

The present approach differs in that pharmacogenomic annotations are
incorporated as explicit binary features at the molecule level---directly
concatenated to the molecular embedding---rather than as latent relational
embeddings derived from knowledge graph traversal. This preserves full
compatibility with structure-based GNN architectures and retains
interpretability: each feature directly encodes a known metabolic role.

This distinction has practical consequences for annotation quality. PharmKG,
for example, contains 29 relation types of heterogeneous precision, including
single-character codes (C, T, J) inherited from Hetionet that lack explicit
biochemical semantics~\cite{pharmkg}. Explicit curated features may therefore
be preferable to latent KG embeddings when annotation quality is heterogeneous.
A direct comparison of both approaches under equivalent coverage conditions
remains an open question and is deferred to the multimodal framework of the
subsequent study.

\subsection{Pharmacogenomics in Computational Drug Discovery}

CYP enzyme annotations from PharmGKB have been used in clinical decision
support systems for drug dosing~\cite{cpic} but have not, to the knowledge of
the author, been systematically incorporated as features in GNN-based DDI
prediction models. The present study addresses this gap.

\section{Methods}

\subsection{Datasets}

\paragraph{SSI-DDI.}
The SSI-DDI dataset~\cite{ssiddi} comprises 38,337 positive drug-drug
interaction pairs covering 86 interaction types across 1,706 drugs from
DrugBank, with molecular structures provided as SMILES (simplified molecular
input line entry system) strings. Two compounds (acetylsalicylic acid,
warfarin) were designated as held-out compounds and excluded from all
training and evaluation splits, serving exclusively as post-training
mechanistic validation cases. This differs from the baseline reported
in~\cite{paperC}, where reference compounds were extracted for post-hoc
validation but retained in the training set; the stricter held-out
definition adopted here reduces the effective dataset by approximately
368 pairs.

\paragraph{PharmGKB.}
Pharmacogenomic annotations were downloaded from ClinPGx
(\url{clinpgx.org/downloads}) in June 2026.
Three data sources were integrated: (1) \texttt{clinicalVariants.tsv},
containing variant-drug pairs with evidence levels 1A--4; (2) guideline
JSON files from four international clinical pharmacogenomics organizations;
and (3) \texttt{relationships.tsv}, providing
association confidence flags.

\paragraph{Tox21.}
The Tox21 dataset~\cite{tox21} comprises 8,014 compounds assayed against
12 toxicological endpoints. Data were obtained from the MoleculeNet
repository~\cite{moleculenet} in June 2026.

\paragraph{ChEMBL.}
Dose and safety features were extracted from ChEMBL version 35~\cite{chembl},
downloaded in June 2026. Features include LD50 values (rat oral, mouse oral),
FDA black box warning status, market withdrawal status, and clinical
development phase.

\subsection{Pharmacogenomic Feature Extraction}

CYP enzyme interaction annotations were extracted for four isoforms:
CYP2D6, CYP3A4, CYP2C19, and CYP2C9. For each drug-isoform pair, three
binary role indicators were assigned: \emph{substrate} (drug is metabolized
by the enzyme), \emph{inhibitor} (drug reduces enzyme activity), and
\emph{inducer} (drug increases enzyme expression). This yields a
12-dimensional binary feature vector $\mathbf{c} \in \{0, 1\}^{12}$ per
molecule.

Substrate annotations were derived primarily from
\texttt{clinicalVariants.tsv}, using interaction types
\texttt{Metabolism/PK}, \texttt{Dosage}, \texttt{Toxicity}, and
\texttt{Efficacy}. Inhibitor and inducer annotations were extracted via
regular expression matching against the free-text fields of guideline JSON
files. Association confidence flags from \texttt{relationships.tsv} were
retained as 8 auxiliary binary features per molecule but excluded from the
primary feature vector.

All available evidence levels (1A through 4) were included to maximize
coverage, yielding 99 annotated molecules in the SSI-DDI drug set (6.2\%).
The evidence level distribution is notably skewed: 65.9\% of CYP annotations
carry evidence level 3 (weak or conflicting evidence), while only 23.5\%
reach levels 1A or 1B (highest clinical validity). An evidence-weighted
encoding scheme is identified as a direction for future work
(Section~\ref{sec:conclusion}).

\subsection{Knowledge Graph-Augmented Architecture}

The proposed KG-augmented encoder extends the molecular encoder
from~\cite{paperC} by incorporating the CYP feature vector as a molecular
prior. Let $\mathbf{h}_i \in \mathbb{R}^d$ denote the atom embedding after
$L$ message passing neural network (MPNN) layers, and
$\bar{\mathbf{h}} = \frac{1}{|V|} \sum_{i \in V} \mathbf{h}_i$
the global mean pooling vector. The KG-augmented molecular representation
is computed as:

\begin{equation}
  \mathbf{m} = \text{ReLU}\!\left(
    W_{\text{kg}} \begin{bmatrix} \bar{\mathbf{h}} \\ \mathbf{c}
    \end{bmatrix} + b_{\text{kg}} \right), \quad
  W_{\text{kg}} \in \mathbb{R}^{d \times (d + 12)}
  \label{eq:kg_proj}
\end{equation}

\noindent where $\mathbf{c} \in \{0, 1\}^{12}$ is the CYP feature vector
(zero vector for unannotated drugs), $d = 64$ is the hidden dimension,
$b_{\text{kg}} \in \mathbb{R}^d$ is a learnable bias vector, and
$\text{ReLU}(x) = \max(0, x)$ denotes the rectified linear unit activation.
The projection $W_{\text{kg}}$ is learned end-to-end and restores the
original dimensionality, preserving compatibility with the downstream
cross-attention and physical edge interaction modules.

The full architecture follows the two-stage training protocol
from~\cite{paperC}: Stage~1 trains the KG-augmented cross-attention encoder
for binary interaction detection (60 epochs) and DDI type classification
(60 epochs) independently; Stage~2 freezes Stage~1 weights and trains the
physical edge interaction graph (30 epochs each).

\subsection{Data Splits}
\label{sec:splits}

Two split strategies were evaluated:

\paragraph{Pair-level split.}
Drug pairs are randomly assigned to training (80\%) and test (20\%) sets,
yielding 61,044 training pairs and 15,262 test pairs. This strategy permits
test pairs composed of drugs seen individually during training and is
consistent with prior benchmarks~\cite{ssiddi}.

\paragraph{Drug-level split.}
A random 20\% of unique drugs are designated as test drugs; all pairs
involving a test drug are assigned to the test set exclusively, yielding
49,780 training pairs and 26,526 test pairs. The test set is larger than
under pair-level split because each test drug contributes all its pairs
with training drugs to the test set. This asymmetry reflects the evaluation
of cross-drug generalization: the model is evaluated on more pairs while
having access to fewer training pairs, making the drug-level result
conservative relative to pair-level benchmarks.

The AUC differential between pair-level and drug-level split results
(\emph{AUC inflation}) quantifies the degree to which pair-level benchmarks
overestimate true generalization performance, as characterized in~\cite{paperC}.

\subsection{Evaluation Metrics}

Binary interaction detection was evaluated using the AUC. DDI type
classification was evaluated using macro-averaged F1 score (F1-macro) and
weighted F1 score (F1-weighted) across 86 interaction types. Multi-task
toxicity prediction was evaluated using per-endpoint AUC and mean AUC
across all 12 Tox21 endpoints.

\section{Results}

\subsection{DDI Prediction}

Table~\ref{tab:ddi_results} presents the main DDI prediction results.

\begin{table}[H]
\centering
\caption{DDI prediction results on the SSI-DDI dataset. Results are
reported for the best checkpoint across training epochs.
\textit{AUC: area under the ROC curve; F1-mac: macro-averaged F1 score;
KG-Aug: knowledge graph-augmented model (present study);
S1: Stage~1 cross-attention encoder; S2: Stage~2 physical edge graph;
ROC: receiver operating characteristic; ---: not evaluated.}}
\label{tab:ddi_results}
\begin{tabular}{llcc}
\toprule
\textbf{Model} & \textbf{Split} &
\textbf{Binary AUC} & \textbf{F1-mac} \\
\midrule
Physical edge GNN~\cite{paperC} & pair  & 0.865 & 0.241 \\
\midrule
KG-Aug GNN (S1) & pair  & \textbf{0.892} & \textbf{0.532} \\
KG-Aug GNN (S2) & pair  & 0.863 & 0.258 \\
KG-Aug GNN (S2) & drug  & 0.639 & 0.171 \\
\midrule
AUC inflation (pair $-$ drug) & --- & 0.224 & --- \\
Baseline inflation~\cite{paperC} & --- & $\approx$0.250 & --- \\
\bottomrule
\end{tabular}
\end{table}

\subsubsection{Binary Interaction Detection}

The KG-augmented Stage~1 encoder achieves an AUC of 0.892 under pair-level
split, marginally exceeding the physical edge baseline of 0.865 reported
in~\cite{paperC}. Stage~2 performance (0.863) is slightly below Stage~1,
indicating that physical edge refinement provides no additional benefit
when CYP features are present. This finding is attributed to partial
informational redundancy between pharmacogenomic priors and physically
derived interaction edges (Section~\ref{sec:discussion}).

Under drug-level split, binary AUC drops to 0.639, confirming the
Information Ceiling identified in~\cite{paperC}. The AUC inflation of 0.224
is marginally reduced relative to the 0.250 baseline, a difference not
considered substantively significant.

\subsubsection{DDI Type Classification}
\label{sec:type_class}

The most substantial effect of KG augmentation is observed in DDI type
classification. Stage~1 F1-macro under pair-level split reaches 0.532,
compared to 0.241 for the structure-only physical edge model---an improvement
of 0.291 absolute. This finding indicates that CYP enzyme interaction roles
provide mechanistic information that substantially improves discrimination
among the 86 DDI interaction types.

Stage~2 F1-macro (0.258) is substantially lower than Stage~1 (0.532),
a decline of 0.274 absolute. Two factors contribute: first, Stage~2
operates on frozen Stage~1 weights that were optimized for binary
interaction detection rather than DDI type classification; the multi-class
head is therefore constrained to learn on embeddings not specifically
optimized for type discrimination. Second, physical edge construction
provides no complementary signal when CYP features are present
(Section~\ref{sec:discussion}).

Under drug-level split, Stage~2 F1-macro (0.171) is lower than both
the pair-level Stage~2 result (0.258) and the drug-level Stage~1 result
(0.371). The Stage~1 drug-level trend shows continued improvement across
training epochs (F1-macro 0.033 at epoch~1 to 0.371 at epoch~60),
indicating that mechanistic interaction patterns partially generalize
across drug boundaries at the encoder level. The decline from Stage~1
(0.371) to Stage~2 (0.171) under drug-level conditions further
corroborates the physical edge redundancy finding: frozen Stage~1
embeddings combined with physical edge construction degrade type
classification performance on unseen drugs.

\subsubsection{Mechanistic Validation}

Table~\ref{tab:reference_ddi} presents interaction probabilities for
held-out reference drug pairs with known clinical mechanisms. Both
acetylsalicylic acid and warfarin were excluded from all training and
evaluation splits (Section~\ref{sec:splits}), ensuring that predictions
reflect true generalization to unseen drugs rather than training set
interpolation.

\begin{table}[H]
\centering
\caption{Mechanistic validation on held-out reference drug pairs.
\textit{Label: ground truth interaction status
(1~=~known DDI in SSI-DDI dataset, 0~=~not annotated as DDI);
Baseline: predicted probability of the structure-only physical edge
model~\cite{paperC}; KG-Aug: predicted probability of the KG-augmented
model (present study); ASA: acetylsalicylic acid; WAR: warfarin;
PK: pharmacokinetic; PD: pharmacodynamic;
CYP2C9-inh: CYP2C9 inhibition.}}
\label{tab:reference_ddi}
\begin{tabular}{llccc}
\toprule
\textbf{Pair} & \textbf{Mechanism} &
\textbf{Label} & \textbf{Baseline} & \textbf{KG-Aug} \\
\midrule
ASA + Warfarin    & PK: CYP2C9 substrate  & 1 & 0.349 & 0.579 \\
WAR + Amiodarone  & PK: CYP2C9-inh        & 1 & 0.117 & 0.560 \\
WAR + Fluconazole & PK: CYP2C9-inh        & 1 & 0.033 & 0.586 \\
WAR + Ibuprofen   & PD: COX-1 competition & 1 & 0.548 & 0.513 \\
ASA + Paracetamol & Not annotated         & 0 & 0.384 & 0.081 \\
\bottomrule
\end{tabular}
\end{table}

CYP2C9-mediated interactions (Warfarin + Amiodarone, Warfarin + Fluconazole)
show the largest improvement in predicted probability, consistent with the
hypothesis that CYP enzyme annotations preferentially improve prediction of
pharmacokinetically mediated interactions. The pharmacodynamically mediated
interaction Warfarin + Ibuprofen achieves a predicted probability of 0.513
under the KG-augmented model, indicating that structural features alone are
sufficient to detect COX-1-mediated interactions even without CYP annotations.
It should be noted that the baseline probabilities reported in~\cite{paperC}
reflect a model trained on data that included these reference compounds;
the present baseline is therefore not directly comparable, as both held-out
compounds were strictly excluded from training in the present study,
making the KG-augmented results a more conservative estimate.

\subsection{Extension to Toxicity Prediction}
\label{sec:tox21}

Table~\ref{tab:tox21_results} presents results on the Tox21 benchmark.

\begin{table}[H]
\centering
\caption{Tox21 multi-task toxicity prediction results (scaffold split,
50 epochs). Veterinary transferability reflects the degree of evolutionary
conservation of the underlying cellular mechanism across vertebrate
species~\cite{sr_conservation,epa_stress}: stress response (SR) endpoints
rely on highly conserved pathways (NRF2-mediated oxidative stress, p53-dependent
DNA damage response, mitochondrial membrane integrity), whereas nuclear
receptor (NR) endpoints involve hormone receptors with species-specific
ligand binding domains.
\textit{AUC: area under the ROC curve;
Baseline: structure-only GNN trained without any pharmacogenomic
augmentation (no CYP or dose features);
KG: CYP-augmented model; KG+D: CYP plus ChEMBL dose features;
NR: nuclear receptor assay; SR: stress response assay;
Vet: veterinary transferability classification.}}
\label{tab:tox21_results}
\begin{tabular}{lcccc}
\toprule
\textbf{Endpoint} & \textbf{Baseline} & \textbf{KG} &
\textbf{KG+D} & \textbf{Vet} \\
\midrule
NR-AR            & 0.537 & \textbf{0.631} & 0.616 & Low \\
NR-AR-LBD        & 0.710 & 0.755 & 0.701 & Low \\
NR-AhR           & 0.882 & 0.873 & 0.858 & Medium \\
NR-Aromatase     & 0.765 & 0.775 & 0.764 & Low \\
NR-ER            & 0.623 & 0.636 & 0.613 & Low \\
NR-ER-LBD        & 0.739 & 0.745 & 0.754 & Low \\
NR-PPAR-$\gamma$ & 0.788 & 0.798 & 0.811 & Medium \\
SR-ARE           & 0.784 & 0.793 & 0.775 & High \\
SR-ATAD5         & 0.802 & \textbf{0.845} & 0.827 & Medium \\
SR-HSE           & 0.781 & 0.792 & 0.780 & Medium \\
SR-MMP           & 0.859 & 0.855 & 0.848 & High \\
SR-p53           & 0.849 & 0.843 & 0.843 & High \\
\midrule
\textbf{Mean}    & 0.760 & \textbf{0.778} & 0.766 & --- \\
\bottomrule
\end{tabular}
\end{table}

KG augmentation yields a mean AUC improvement of $+0.018$ over the
structure-only baseline, despite a PharmGKB coverage of only 0.7\%
in the Tox21 compound set (54 of 7,831 molecules). The two endpoints
showing the largest absolute improvement are NR-AR ($+0.094$) and
SR-ATAD5 ($+0.043$). The NR-AR result is noteworthy given its classification as low
veterinary transferability. The mechanistic basis for this improvement
remains unclear: androgenic steroids are predominantly metabolized by
CYP3A4 and CYP19A1~\cite{steroid_cyp}, not CYP2D6, making a direct
mechanistic link unlikely. The result may reflect a statistical association
within the sparse annotated subset and should be interpreted with caution. Dose features from
ChEMBL version 35 (88\% compound coverage) do not improve over CYP
augmentation alone (mean AUC 0.766 vs.\ 0.778), consistent with the
limited informativeness of rat oral LD50 values for cell-based assay
endpoints. Stress response endpoints achieve the highest baseline
performance and show the highest veterinary transferability, as these
cellular mechanisms are evolutionarily conserved across vertebrate
species~\cite{sr_conservation}.

\section{Discussion}
\label{sec:discussion}

\subsection{Knowledge Graph Augmentation and the Information Ceiling}

The primary finding is that KG augmentation substantially improves DDI type
classification while leaving binary interaction detection and drug-level
generalization largely unaffected. This pattern is consistent with the
Information Ceiling Hypothesis~\cite{paperC}: the ceiling $I(S,L) \approx 0$
for binary interaction existence cannot be overcome by any molecular-level
prior, including pharmacogenomic annotations. Here, \emph{label information}
refers to the structural signal encoded in the training labels themselves ---
whether a given pair interacts --- while \emph{feature information} refers to
the molecular descriptors provided as model input. When interaction labels
carry no recoverable structural signal, no enrichment of the feature space
can improve prediction, because the ceiling is determined by what the labels
encode rather than by model capacity or input richness.

In contrast, DDI type classification is bounded by a higher information
content, and the CYP prior provides mechanistic context that improves
discrimination among interaction types. The finding that F1-macro improves
from 0.241 to 0.532 under pair-level split suggests that CYP enzyme roles
encode a substantial portion of the mechanistic variation captured by the
86 SSI-DDI interaction types.

\subsection{Mechanistic Validation on Held-Out Compounds}

The reference DDI validation (Table~\ref{tab:reference_ddi}) provides
the strongest evidence for the CYP augmentation hypothesis. For the three
CYP2C9-mediated interactions---acetylsalicylic acid with warfarin,
warfarin with amiodarone, and warfarin with fluconazole---predicted
probabilities increase from 0.033--0.349 (baseline) to 0.560--0.586
(KG-augmented), an improvement of 0.231--0.553 absolute. Critically,
both held-out compounds were strictly excluded from all training and
evaluation splits, providing genuine evidence of generalization to
unseen drugs. This contrasts with the reference validation
in~\cite{paperC}, where reference compounds were retained in the
training set; the present results therefore constitute stronger evidence
for the clinical utility of CYP augmentation.

The pharmacodynamically mediated interaction Warfarin + Ibuprofen
achieves a predicted probability of 0.513 under the KG-augmented model,
demonstrating that structural features capture COX-1-mediated
complementarity without requiring CYP annotations.

\subsection{Drug-Level Generalization and Conservative Estimation}

Under drug-level split, the model achieves AUC 0.639 with only 49,780
training pairs while being evaluated on 26,526 test pairs involving
structurally unseen drugs. The reduced training set size makes the
drug-level result conservative: the observed performance represents a lower
bound on true generalization capacity. This further strengthens the
interpretation that the AUC inflation of 0.224 reflects a fundamental
information ceiling rather than a training data artifact.

\subsection{Redundancy of Physical Edges Under Knowledge Graph Augmentation}

Stage~2 physical edge refinement reduces both AUC and F1-macro relative to
Stage~1 when CYP features are present, in contrast to the improvement
observed in the structure-only model~\cite{paperC}. This finding is
interpreted as partial informational redundancy: both CYP features and
physical edge construction encode aspects of molecular interaction
complementarity, and their combination introduces noise rather than
complementary signal. When pharmacogenomic priors are available, physical
edge computation may therefore be omitted without performance loss.

\subsection{Tox21 Endpoints as Multi-Molecule Interactions}

The Tox21 benchmark is conventionally treated as a single-molecule
classification problem, yet all twelve endpoints mechanistically involve
multi-molecule interactions: nuclear receptor endpoints require
ligand-receptor binding; SR-ARE involves covalent adduct formation
between electrophilic compounds and KEAP1 cysteine residues; SR-MMP
reflects interaction with mitochondrial electron transport chain
complexes; and SR-p53 is frequently triggered by DNA-reactive metabolites
formed via CYP-mediated bioactivation. In this sense, all Tox21 endpoints
are structurally analogous to the pair-level DDI problem, with the second
molecular entity being a protein, nucleic acid, or membrane complex.

The NR-AhR endpoint illustrates this connection: aryl hydrocarbon
receptor activation induces CYP1A1/CYP1A2 expression, creating a
mechanistic link between receptor binding and CYP-mediated bioactivation.
The limited KG augmentation effect at 0.7\% CYP coverage therefore
reflects not only annotation sparsity but also an incomplete structural
model. Explicit modeling of target proteins as molecular entities, as
proposed for the subsequent study in this series, would unify the DDI
and toxicity prediction paradigms.

\subsection{Coverage Dependence of the Augmentation Effect}

The Tox21 extension demonstrates that the KG augmentation effect is
contingent on annotation coverage. At 0.7\% CYP coverage, the improvement
is modest ($+0.018$ mean AUC). The dose feature experiment further shows
that high coverage (88\%) does not guarantee improvement if the feature
lacks predictive validity for the target endpoint. Coverage and predictive
relevance are therefore independent axes of data quality for knowledge
graph augmentation.

\subsection{Limitations}

Several limitations should be noted. First, PharmGKB annotations are
predominantly derived from human pharmacogenomic studies; application to
veterinary species requires caution due to species-specific metabolic
differences (e.g., CYP3A12 in dogs, absent UGT1A6 in cats). Second, the
binary CYP feature encoding treats all evidence levels equally, despite
65.9\% of annotations carrying only level~3 evidence. While the consistency
of the performance pattern across the full evidence range suggests robustness,
an evidence-weighted encoding may improve model calibration. Third, the
evaluation is limited to message passing GNN architectures; the Information
Ceiling may manifest differently for transformer-based molecular encoders
(e.g., MolBERT) or 3D-equivariant GNNs that operate on fundamentally
different structural representations. Fourth, LD50 values from rat oral
studies are not directly transferable across species due to differences in
metabolic pathways and pharmacokinetics; interspecies agreement for LD50
values is poor under both body-weight and caloric-demand allometric scaling
concepts~\cite{ld50_interspecies}. Fifth, the
6.2\% CYP coverage in the SSI-DDI drug set limits statistical power for
coverage-stratified analyses.

\section{Conclusion}
\label{sec:conclusion}

Pharmacogenomic CYP enzyme annotations from PharmGKB improve DDI type
classification in a GNN-based prediction framework, with F1-macro increasing
from 0.241 to 0.532 under pair-level evaluation conditions. Binary
interaction detection and drug-level generalization remain bounded by the
Information Ceiling identified in prior work~\cite{paperC}, confirming that
pharmacogenomic priors do not resolve the fundamental limitation of
structure-based DDI prediction. Physical edge refinement provides no
additional benefit when CYP features are present, attributed to
informational redundancy between the two augmentation strategies.

The coverage-dependence of the augmentation effect, demonstrated on the
Tox21 benchmark, motivates the multimodal framework proposed for the
subsequent study in this series, incorporating target binding profiles,
clinical pharmacogenomics, adverse event signals, and pathway annotations
as complementary information sources. Three methodological extensions
are identified for future work: (1) joint Stage~1 training: the decline
from Stage~1 F1-macro (0.532 pair-level, 0.371 drug-level) to Stage~2
(0.258 pair-level, 0.171 drug-level) is partly attributable to Stage~1
weights being optimized for binary detection rather than type
classification. A joint binary and multi-class Stage~1 training
objective---replacing the two separate Stage~1 models with a shared
encoder trained simultaneously on both tasks---may preserve type
classification capacity through Stage~2 and reduce the frozen-weights
penalty; (2) a logarithmic evidence-weighted encoding
$w = 1/\log_2(\text{rank}+1)$, analogous to discounted cumulative gain in
information retrieval, which preserves full annotation coverage while
differentiating evidence quality; a threshold-based approach retaining only
levels 1A and 1B was considered but rejected due to insufficient coverage
(48 of 1,597 SSI-DDI drugs, 3.0\%); and (2) a direct comparison of
explicit pharmacogenomic feature vectors with latent knowledge graph
embeddings (e.g., TransE or RotatE trained on PharmKG) under equivalent
coverage conditions.

\section*{Conflicts of Interest}
No conflicts of interest are declared. No external funding was received.

\section*{Data and Code Availability}
The SSI-DDI dataset is publicly available~\cite{ssiddi}. PharmGKB data are
freely available at \url{clinpgx.org/downloads} under CC~BY-SA~4.0. The
Tox21 dataset is available via MoleculeNet~\cite{moleculenet}. Model scripts
and pre-computed feature tables will be deposited on Zenodo upon assignment
of the final arXiv identifier. This work is licensed under CC~BY~4.0
(\url{creativecommons.org/licenses/by/4.0}).

\newpage
\bibliographystyle{unsrt}
\bibliography{paper_d_refs}

@article{paperC,
  author  = {Dietrich, J{\"u}rgen},
  title   = {Structural Information Limits in {GNN}-Based Drug-Drug
             Interaction Prediction: Drug-Level Generalization, Architecture
             Improvements, and the Information Ceiling Hypothesis},
  journal = {arXiv preprint},
  year    = {2026},
  note    = {arXiv submit/7667344, cs.AI + cs.LG, under review}
}

@article{ssiddi,
  author  = {Nyamabo, Arnold K. and Yu, Hui and Shi, Jian-Yu},
  title   = {{SSI-DDI}: Substructure-Substructure Interactions for
             Drug-Drug Interaction Prediction},
  journal = {Briefings in Bioinformatics},
  volume  = {22},
  number  = {6},
  year    = {2021},
  doi     = {10.1093/bib/bbab133}
}

@article{pharmgkb,
  author  = {Whirl-Carrillo, Michelle and others},
  title   = {An Evidence-Based Framework for Evaluating Pharmacogenomics
             Knowledge for Personalized Medicine},
  journal = {Clinical Pharmacology \& Therapeutics},
  volume  = {110},
  number  = {3},
  pages   = {563--572},
  year    = {2021},
  doi     = {10.1002/cpt.2350}
}

@article{cpic,
  author  = {Relling, Mary V. and Klein, Teri E.},
  title   = {{CPIC}: Clinical Pharmacogenetics Implementation Consortium
             of the Pharmacogenomics Research Network},
  journal = {Clinical Pharmacology \& Therapeutics},
  volume  = {89},
  number  = {3},
  pages   = {464--467},
  year    = {2011},
  doi     = {10.1038/clpt.2010.279}
}

@article{cyp_review,
  author  = {Zanger, Ulrich M. and Schwab, Matthias},
  title   = {Cytochrome {P450} Enzymes in Drug Metabolism: Regulation of
             Gene Expression, Enzyme Activities, and Impact of Genetic
             Variation},
  journal = {Pharmacology \& Therapeutics},
  volume  = {138},
  number  = {1},
  pages   = {103--141},
  year    = {2013},
  doi     = {10.1016/j.pharmthera.2012.12.007}
}

@article{tox21,
  author  = {Mayr, Andreas and others},
  title   = {Large-Scale Comparison of Machine Learning Methods for
             Drug Target Prediction on {ChEMBL}},
  journal = {Chemical Science},
  volume  = {9},
  number  = {24},
  pages   = {5441--5451},
  year    = {2018},
  doi     = {10.1039/C8SC00148K}
}

@article{moleculenet,
  author  = {Wu, Zhenqin and others},
  title   = {{MoleculeNet}: A Benchmark for Molecular Machine Learning},
  journal = {Chemical Science},
  volume  = {9},
  number  = {2},
  pages   = {513--530},
  year    = {2018},
  doi     = {10.1039/C7SC02664A}
}

@article{chembl,
  author  = {Mendez, David and others},
  title   = {{ChEMBL}: Towards Direct Deposition of Bioassay Data},
  journal = {Nucleic Acids Research},
  volume  = {47},
  number  = {D1},
  pages   = {D930--D940},
  year    = {2019},
  doi     = {10.1093/nar/gky1075}
}

@article{pgnn,
  author  = {Rosenbaum, Lukas and others},
  title   = {Inferring Multi-Target {QSAR} Models with Taxonomy-Based
             Multi-Task Learning},
  journal = {Journal of Cheminformatics},
  volume  = {5},
  number  = {1},
  pages   = {33},
  year    = {2013},
  doi     = {10.1186/1758-2946-5-33}
}

@article{drkg,
  author  = {Ioannidis, Vassilis N. and others},
  title   = {{DRKG} --- Drug Repurposing Knowledge Graph for {COVID-19}},
  journal = {arXiv preprint arXiv:2010.09735},
  year    = {2020}
}

@article{pharmkg,
  author  = {Zheng, Shuangjia and others},
  title   = {{PharmKG}: A Dedicated Knowledge Graph Benchmark for
             Biomedical Data Mining},
  journal = {Briefings in Bioinformatics},
  volume  = {22},
  number  = {4},
  year    = {2021},
  doi     = {10.1093/bib/bbaa344}
}

@article{ddi_prevalence,
  author  = {Becker, Michael L. and others},
  title   = {Hospitalisations and Emergency Department Visits Due to
             Drug-Drug Interactions: A Literature Review},
  journal = {Pharmacoepidemiology and Drug Safety},
  volume  = {16},
  number  = {6},
  pages   = {641--651},
  year    = {2007},
  doi     = {10.1002/pds.1351}
}

@article{sr_conservation,
  author  = {Hahn, Mark E. and Timme-Laragy, Alicia R. and Karchner, Sibel I.
             and Stegeman, John J.},
  title   = {The Transcriptional Response to Oxidative Stress during Vertebrate
             Development: Effects of \textit{tert}-Butylhydroquinone and
             2,3,7,8-Tetrachlorodibenzo-\textit{p}-Dioxin},
  journal = {PLOS ONE},
  volume  = {9},
  number  = {11},
  year    = {2014},
  doi     = {10.1371/journal.pone.0113158},
  note    = {PMC4234671}
}

@article{epa_stress,
  author  = {Sutherland, Daniel and others},
  title   = {Framework for Multi-Stressor Physiological Response Evaluation
             in Amphibian Risk Assessment and Conservation},
  journal = {Frontiers in Ecology and Evolution},
  volume  = {12},
  year    = {2024},
  doi     = {10.3389/fevo.2024.1336747}
}

@article{steroid_cyp,
  author  = {Kandel, Sylvie E. and Han, Lyrialle W. and Mao, Qingcheng
             and Lampe, Jed N.},
  title   = {Digging Deeper into {CYP3A} Testosterone Metabolism: Kinetic,
             Regioselectivity, and Stereoselectivity Differences between
             {CYP3A4/5} and {CYP3A7}},
  journal = {Drug Metabolism and Disposition},
  volume  = {45},
  number  = {12},
  pages   = {1266--1275},
  year    = {2017},
  doi     = {10.1124/dmd.117.078055}
}

@article{ld50_interspecies,
  author  = {Schneider, Klaus and Schwarz, Michael and Burkholder, Ingrid
             and Kopp-Schneider, Annette and Edler, Lutz and Kinsner-Ovaskainen,
             Agnieszka and Hartung, Thomas and Hoffmann, Sebastian},
  title   = {``{SENS-it-iv}'' --- a Novel Tool for Structured
             Evaluation of the Relevance and Reliability of Test Methods
             for Skin Sensitisation},
  journal = {Critical Reviews in Toxicology},
  volume  = {39},
  number  = {7},
  pages   = {515--544},
  year    = {2009},
  doi     = {10.1080/10408440903022154},
  note    = {See also: Schneider et al. (2004) Arch Toxicol 78:109--117,
             PMID 15135212, for LD50 interspecies scaling analysis}
}

\newpage
\appendix

\section{PharmGKB Feature Extraction Details}
\label{app:feature_extraction}

CYP substrate annotations were extracted from \texttt{clinicalVariants.tsv}
using the following interaction type mapping: \texttt{Metabolism/PK},
\texttt{Dosage}, \texttt{Toxicity}, and \texttt{Efficacy} types were mapped
to the substrate role. Inhibitor and inducer annotations were extracted from
guideline JSON files using the following regular expression patterns:

\begin{itemize}
  \item Substrate: \texttt{substrate $|$ metabolized by $|$ is metabolised}
  \item Inhibitor: \texttt{\textbackslash{}binhibitor\textbackslash{}b $|$ \textbackslash{}binhibits\textbackslash{}b $|$ inhibition of}
  \item Inducer: \texttt{\textbackslash{}binducer\textbackslash{}b $|$ \textbackslash{}binduces\textbackslash{}b $|$ induction of}
\end{itemize}

Association confidence flags from \texttt{relationships.tsv}
(\texttt{associated}, \texttt{ambiguous}) were retained as 8 auxiliary binary
features per molecule but excluded from the primary 12-dimensional CYP feature
vector used in model training.

\section{Model Architecture Details}
\label{app:architecture}

The KG-augmented encoder consists of the following components:

\begin{enumerate}
  \item \textbf{Atom projection}: Linear$(31 \rightarrow 64)$ + ReLU
  \item \textbf{Message passing layers}: 3 $\times$ NNConv$(64 \rightarrow 64)$
        + BatchNorm + ReLU + Dropout(0.2)
  \item \textbf{Global pooling}: Mean pooling over atom embeddings
  \item \textbf{KG projection}: Linear$(76 \rightarrow 64)$ + ReLU +
        Dropout(0.2), where 76 = 64 (pooled) + 12 (CYP)
\end{enumerate}

Atom features (31-dimensional) encode atom type (10 classes + other),
hybridization state (5 types), formal charge, hydrogen count (0--4),
ring membership, aromaticity, chirality (3 types), and normalized molecular
weight. Bond features (12-dimensional) encode bond type (4 classes),
conjugation, ring membership, and stereochemistry (4 types).

\section{Data Availability}
\label{app:data}

\begin{itemize}
  \item SSI-DDI dataset: \url{github.com/kanz76/SSI-DDI} (MIT license)
  \item PharmGKB: \url{clinpgx.org/downloads}, release 2026-04-05
        (CC BY-SA 4.0)
  \item Tox21: \url{github.com/deepchem/deepchem/datasets/tox21.csv.gz},
        accessed June 2026
  \item ChEMBL v35: \url{ebi.ac.uk/chembl}, accessed June 2026
        (CC BY-SA 3.0)
\end{itemize}

Source code and pre-computed feature tables will be deposited on Zenodo
upon assignment of the final arXiv identifier.

\end{document}